%% file: iclr2025_conference.tex
\documentclass{article} 
\usepackage{iclr2025_conference,times}

\usepackage[utf8]{inputenc} 
\usepackage[T1]{fontenc}    
\usepackage{url}            
\usepackage{booktabs}       
\usepackage{amsfonts}       
\usepackage{nicefrac}       
\usepackage{microtype}      
\usepackage{xcolor}         

\usepackage{amssymb}
\usepackage{bbm}
\newcommand{\blue}[1]{\textcolor{blue}{#1}}

\usepackage{enumitem}

\usepackage{wrapfig}

\usepackage{booktabs}

\input{math_commands.tex}

\usepackage{graphicx}
\graphicspath{{assets/}} 

\usepackage{lipsum}
\usepackage{makecell} 
\usepackage{subfigure}
\usepackage[pagebackref,breaklinks,colorlinks]{hyperref}
\usepackage{multirow}

\usepackage{colortbl}
\usepackage{pifont} 
\newcommand{\cmark}{\textcolor{green}{\ding{51}}} 
\newcommand{\xmark}{\textcolor{red}{\ding{55}}}   

\usepackage{xspace}
\makeatletter
\DeclareRobustCommand\onedot{\futurelet\@let@token\@onedot}
\def\@onedot{\ifx\@let@token.\else.\null\fi\xspace}

\def\eg{\emph{e.g}\onedot} 
\def\ie{\emph{i.e}\onedot}

\makeatother

\usepackage[pagebackref,breaklinks,colorlinks]{hyperref}
\hypersetup{
	colorlinks=true,
	linkcolor=red,
	filecolor=blue,      
	urlcolor=cyan,
	citecolor=green,
}

\usepackage{hyperref}
\usepackage{url}

\title{Towards Hierarchical Multi-Agent Workflows for Zero-Shot Prompt Optimization}

\author{Yuchi Liu$^1$, Jaskirat Singh$^1$, Gaowen Liu$^2$, Ali Payani$^2$, Liang Zheng$^1$ \\
$^1$Australian National University, $^2$Cisco \\
\texttt{\{yuchi.liu, jaskirat.sing, liang.zheng\}@anu.edu.au}\\
\texttt{\{gaowen.liu, ali.payani\}@cisco.com} \\
}

\iclrfinalcopy 
\begin{document}

\maketitle
\begin{abstract}
  Large language models (LLMs) have shown great progress in responding to user questions, allowing for a multitude of diverse applications. Yet, the quality of LLM outputs heavily depends on the prompt design, where a good prompt might enable the LLM to answer a very challenging question correctly. Therefore, recent works developed many strategies for improving the prompt, including both manual crafting and in-domain optimization. 
  However, their efficacy in unrestricted scenarios remains questionable, as the former depends on human design for specific questions and the latter usually generalizes poorly to unseen scenarios. 
  To address these problems, we give LLMs the freedom to design the best prompts according to themselves. Specifically, we include a hierarchy of LLMs, first constructing a prompt with precise instructions and accurate wording in a hierarchical manner, and then using this prompt to generate the final answer to the user query. We term this pipeline Hierarchical Multi-Agent Workflow, or HMAW. 
  In contrast with prior works, HMAW imposes no human restriction and requires no training, and is  
  completely task-agnostic while capable of adjusting to the nuances of the underlying task.
  Through both quantitative and qualitative experiments across multiple benchmarks, we verify that despite its simplicity, the proposed approach can create detailed and suitable prompts, further boosting the performance of current LLMs. Project page: \href{https://liuyvchi.github.io/HMAW_project/}{liuyvchi.github.io/HMAW\_project/}
\end{abstract}

\section{Introduction}




Large language models (LLMs) can perform a wide range of tasks such as creating detailed literary works and generating computer codes. 
Yet they often give unsatisfactory responses when the prompt is poorly designed. 
To design better prompts that make the LLM more effective, prompt optimization has been extensively explored in recent years. Specifically, existing works on prompt optimization can be categorized as follows.
The first category involves the manual design of prompts, such as the chain of thought (CoT) \citep{wei2022chain} and its variants \citep{kojima2022large, besta2023graph, yao2023tree}. 
The second category focuses on optimizing prompts on a training set that contains input-response pairs to learn a golden prompt, as demonstrated by APE \citep{zhou2023large}. 
The third category also uses handcrafted prompts, but different from the first category, the handcrafted prompts include examples that can guide an LLM agent to further generate prompts for a second LLM that actually provide the answer, \eg, ExpertPrompting \citep{xu2023expertprompting}.

\begin{figure}[t]
\vskip -0.5cm
\begin{center}
\centerline{\includegraphics[width=1.\linewidth]{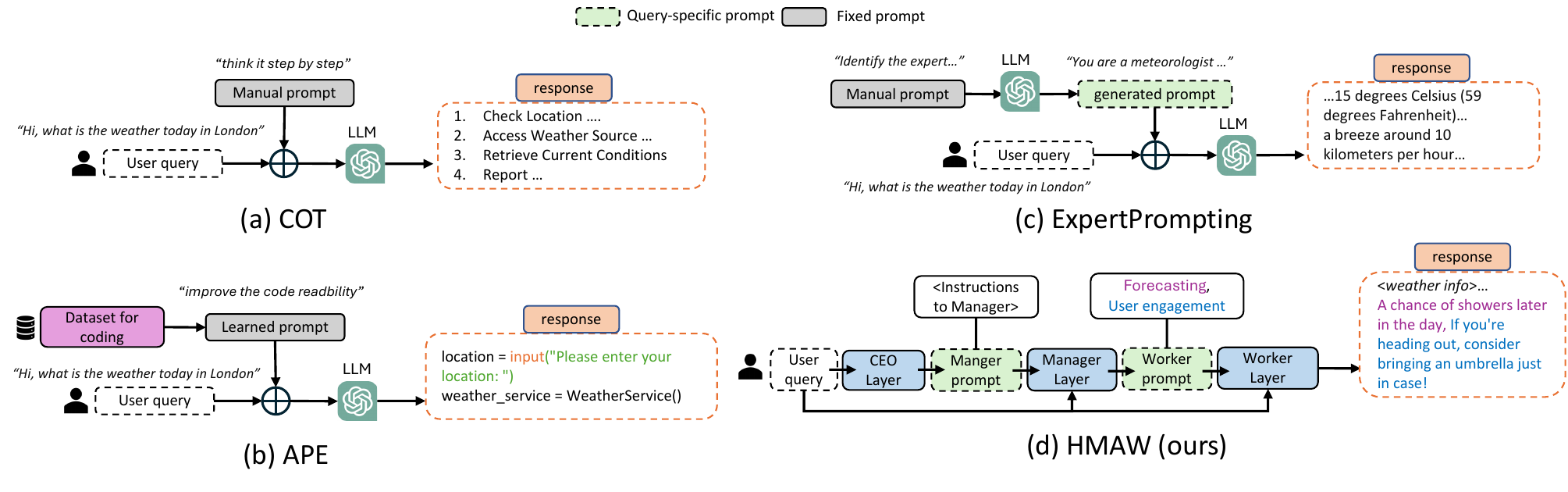}}
\vskip -0.1in
\caption{\textbf{Examples comparing the generalization ability of existing methods and the proposed one.} 
(a) COT \citep{wei2022chain} uses a handcrafted prompt, which might not be suitable for all tasks.
(b) APE \citep{zhou2023large} fine-tunes the prompt on a specific dataset, and its generalization capability to other scenarios is questionable. 
(c) ExperPrompting \citep{xu2023expertprompting} includes few-shot examples in the system prompt to help an LLM convert the user query to a format more suitable for LLM, but these examples might not be able to cover all scenarios. 
(d) Our method adopts a hierarchical design in reformatting the user query. Free from pre-defined few-shot examples, the interaction between the hierarchy allows for more generalizable yet more adaptive tuning of prompts.  
}
\label{fig:problem-overview}
\end{center}
\vskip -0.1in
\end{figure}

However, these methods generally have limited generalization performance across various tasks, as shown in Fig. \ref{fig:problem-overview} (a), (b), and (c). 
The first category \citep{wei2022chain} relies on \textit{handcrafted prompts}. While they are very useful for tasks such as math, the same prompting method, if applied to other tasks, would be less effective. 
The second category, \citep{zhou2023large,yang2024large}, \textit{fine-tuned} on a specific dataset, can obtain good performance for in-domain tasks. However, these learned prompts from one domain might fail for other tasks, making them not task-agnostic. 
The third category \citep{zhou2023large}, which \textit{uses LLM to adjust the prompt}, is still limited by the pre-defined examples in the handcrafted prompts. These examples only cover a finite number of scenarios and can potentially limit the quality of the modified prompt.

In this paper, we introduce a more generalizable prompt optimization method named hierarchical multi-agent workflow (HMAW). In a nutshell, our method mimics the hierarchy in companies, where the CEO and the manager create the guidelines for workers, and workers execute user-specified tasks while following these guidelines. Similarly, our method has a CEO $\rightarrow$ Manager $\rightarrow$ Worker workflow, which takes the initial user query as input and outputs the user response. 
The CEO and the Manager LLMs work in a hierarchical manner to generate accurate and detailed instructions, which are then used as prompts for the Worker LLM to generate the final answer. 

Compared with existing prompt optimization methods, HMAW has two advantages that contribute to its generalization ability. 
First, having a hierarchy of LLM agents that each has its own job greatly simplifies the task for them. In this way, LLMs can either focus on either the overall goal (CEO), or creating a more detailed checklist (Manager), or giving specific answers (Worker). 
Second, our method does not require few-shot examples or any training set, thus does not fit to certain tasks, and has the additional benefit to be easy to use. As such, our method is zero-shot, task-agnostic, and prompt-specific. We compare HMAW with  existing literature in Table \ref{tab:method_compare}. 

On multiple benchmarks, we experimentally verify the effectiveness of the proposed hierarchical approach. When combined with Mixtral \citep{jiang2024mixtral}, HMAW achieves a significant average improvement of 30.7\% across 5 datasets.

\section{Related Work}\label{sec:related-work}

\textbf{Manual prompt engineering} integrates human-like problem-solving knowledge. 
For example, few-shot prompting \citep{brown2020language} allows language models (LMs) to generate responses by providing them with explicit examples. The chain-of-thought (CoT) method \citep{wei2022chain}, along with its variants such as zero-shot CoT \citep{kojima2022large}, graph-of-thought (GoT) \citep{besta2023graph}, and tree-of-thought (ToT) \citep{yao2023tree}, intricately designs prompts to emulate various types of human-like reasoning processes. Some other works \citep{welleck2023generating, shinn2023reflexion, gou2024critic} manually craft prompts that encourage LLMs to engage in critical thinking and verification processes before delivering the final answer. Those manual prompt engineering techniques typically require carefully tailored prompts for various tasks, and for each task, the prompts used remain the same across different user queries. In comparison, we design a task-agnostic prompter that can create tailored prompts for every user query.

\begin{table}[t]
\centering
\small
\setlength{\tabcolsep}{4mm}{
\begin{tabular}{l|c|c|c}
\toprule
Method & Zero-Shot & Task-agnostic  &  Query-Specific   \\ 
\midrule
Chain of Thought (CoT) \ \cite{wei2022chain} & \xmark & \cmark & \xmark \\
Zero-shot CoT \ \cite{kojima2022large} & \cmark & \cmark & \xmark \\
Static ExpertPromting \citep{xu2023expertprompting} & \cmark & \cmark & \xmark \\
Dynamic ExpertPromting \citep{xu2023expertprompting} & \xmark & \xmark & \cmark \\
on Meta-Prompting \citep{de2023meta} & \xmark & \cmark & \cmark \\
Rephrase and Response \citep{deng2023rephrase} & \cmark & \cmark & \xmark \\
Multi-Persona \citep{wang2023unleashing}  & \xmark & \cmark & \cmark \\ 
APE \citep{zhou2023large}  & \xmark & \xmark & \xmark \\ 
PromptAgent \citep{wang2024promptagent}  & \xmark & \xmark & \xmark \\ 
HMAW (Ours) & \cmark & \cmark & \cmark \\ 
\bottomrule
\end{tabular}
}
\caption{Method comparison based on various properties. Our method is zero-shot, task-agnostic, and query-specific, which are useful for dealing with diverse tasks.}
\label{tab:method_compare}
\end{table}

    


\textbf{Automated prompt engineering methods.} Instead of using fixed hand crafted in-context prompts, automated prompt engineering methods are attracting increasing attention. This line of research aims to let LLMs be prompt generators which automatically generate specific prompts for each user query so that LLMs can respond better. Recent methods include automatic prompt engineer (APE) \citep{zhou2023large}, expert prompting (EP) \citep{xu2023expertprompting},  synthetic prompting \citep{shao2023synthetic}, skeleton-of-thought (SoT), \citep{ning2024skeletonofthought}, multi-personas \citep{wang2023unleashing}, on-meta-prompting \cite{de2023meta},  meta-prompting \citep{suzgun2024meta}, \textit{etc}. While these methods can generate specific prompts according to different original prompts or instructions, they are generally not task-agnostic or  zero-shot, where manually designed examples are required. In other words, they would not be very suitable for generalizing across ever changing landscape of LLMs. In comparison, our method is completely zero-shot, task-agnostic and prompt-specific. 

%

\textbf{Optimization based self-prompt refinement.} Using optimization techniques to improve prompts is a recent focus. Early attempts including prompt tuning \cite{Lester2021ThePO} and prefix tuning \cite{Li2021PrefixTuningOC} tried to learn a trainable prefix or soft prompts for better prompting. Recent ones, \textit{e.g.,}, directional stimulus prompting (DSP) \citep{li2024guiding}, prompt-OIRL \cite{sun2023query}, use reinforcement learning to train an auxiliary model to generate query-specific prompts. OPRO \citep{yang2024large} makes LLMs itself a optimizer which guides LLMs to produce outputs that align with a predefined objective. 
Those works typically require numerous iterations of training for each task and are thus not task-agnostic, and it is costly to adapt them to new tasks. Our work also optimizes user prompt and is thus closely related to this line of works. However, our method is learning-free and task-agnostic.







\textbf{Zero-shot self-prompt refinement.} Some manual prompt engineering methods, such as Zero-shot CoT \citep{kojima2022large}, Static ExpertPrompting \citep{xu2023expertprompting}, and RaR \citep{deng2023rephrase}, follow the zero-shot paradigm. Because no prompt examples or datasets are used, these methods are typically task-agnostic as well. Despite their simplicity and ease of adaptation to various tasks, their prompt optimization capability is limited by their shallow understanding of the user query. More importantly, they are not guaranteed to be query-specific (e.g., Static ExpertPrompting and RaR), which further limits their capability. In this work, we design a hierarchical multi-agent workflow to address these shortcomings.

\section{Proposed Approach}

\subsection{Preliminaries}

\textbf{Notations.}
In language generation tasks, the language model $\mathcal{M}$ is asked to provide responses to queries in the textual space $\mathcal{T}$. Given a dataset $\mathcal{D} = \{ (q_i, y_i^*) \}_{i=1}^n$, where $n$ is the number of pairs, 
each $(q_i, y_i^*)$ pair consists of a text query $q_i \in \mathcal{T}$ and the corresponding golden response $y_i^* \in \mathcal{T}$. 
We use prompt $p \in \mathcal{T}$ to denote input of the large language model (LLM) $\mathcal{M}$. In practice, $p$ may not be equal to $q$ and can be the prompt engineering result $f(q)$, where $f:\mathcal{T}\rightarrow \mathcal{T}, f \in \mathcal{F}$, is the prompt-engineering mapping function. In evaluation,
 we use a metric $s : \mathcal{T} \times \mathcal{T} \rightarrow \mathbb{R}$ to measure the quality of the responses $\hat{y}_i = \mathcal{M}(f(q_i))$.
 If $\mathcal{D}$ provides golden responses, $s$ measures the correctness of $y_i$, \ie, $s(\hat{y_i}, y_i^*) = \mathbbm{1}\{\hat{y_i} = y_i^*\}$, \emph{e.g.}, when $\mathcal{D}$ consists of mathematical or multiple choice questions where answers are objective. If $\mathcal{D}$ involves open questions whose golden responses are not predetermined, we use $s(\hat{y}_i)$ to measure the quality of the response $\hat{y}_i$ generated by $\mathcal{M}$.

\textbf{Prompt optimization.} Directly prompting LLMs $\mathcal{M}$ using the initial human query $q$ may not be the optimal. Ideally, an optimal prompt $p*$ allows $\mathcal{M}$ to generate the golden response $y^*$. Given a specific pair $(q_i, y_i^*)$, the task of finding such $p_i^*$ can be formulated as a prompt optimization task which aims to maximize the following expected quality score:
\begin{equation} 
p_i^* = \underset{p\in \mathcal{T}}{\arg\max} \,  s( y_i^*, \mathcal{M}(p\mid q_i)).
\label{eq:prompt_opt}
\end{equation}

Some methods 
manually combine $q_i$ with some carefully designed fixed prefix or suffix to formulate $p_i^*$. For example, few-shot prompting \citep{brown2020language} uses a few examples of query-answer pairs beside the query of interest $q_i$; the zero-shot chain-of-thought (CoT) \citep{wei2022chain} is formulated as `$q_i\oplus [\text{let us think step by step}]$'. Others use automatic methods, which are described in Section \ref{sec:related-work}.  

A recent line of research (\eg, RAR \citep{deng2023rephrase}, ExpertPrompting \citep{xu2023expertprompting}, Multi-Persona \citep{wang2023unleashing}) focuses on leveraging the text generation capacity of LLMs to directly generate $p_i^*$ by providing some context $C \in \mathcal{T}$ of the optimization task in Eq. \ref{eq:prompt_opt}. Those methods are deployment-friendly because of characteristics like zero-shot, task-agnostic, and query-specific.
In this paper, we not only keep those characteristics but also use a workflow to optimize prompts.

\subsection{Hierarchical Multi-Agent Workflow}

\textbf{Intuition.} Now, we consider prompt optimization as a complex and challenging task where prompts are hard to optimize by traditional optimization method. Our design is motivated by recent multi-agent workflow methods \citep{yang2024doraemongpt, wu2024agentkit} which have shown promise in solving complex tasks. We thus employ the idea of cooperative multiple agents in the prompt optimization task. 

\textbf{Hierarchical workflow.}
We design a hierarchical workflow which performs prompt optimization and then task response generation. The user query $q_i$ is the input of the workflow and is processed and improved by a hierarchy of agents. The workflow output the final is task response $\hat{y}_i$ from an LLM prompted by the improved user query.  

\begin{figure}[t]
\vskip -0.5cm
\begin{center}
\centerline{\includegraphics[width=1.\linewidth]{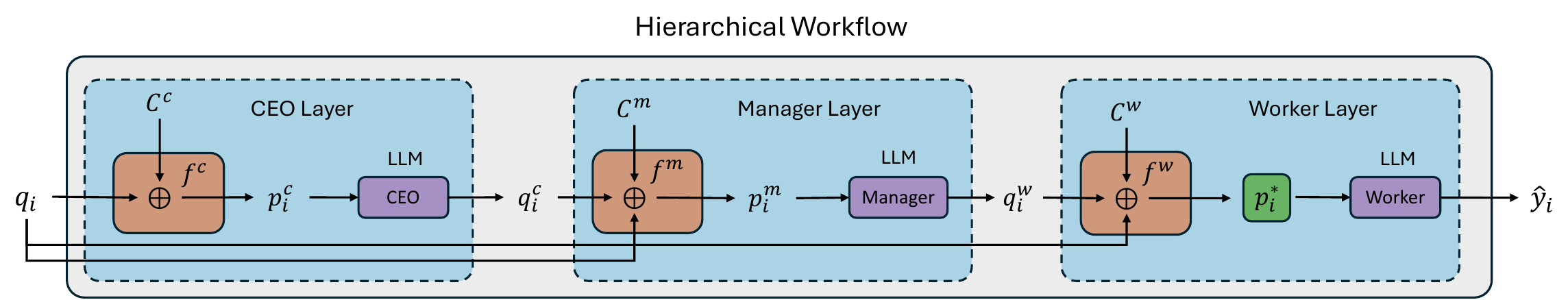}}
\caption{\textbf{Method Overview.} We propose modeling the prompt optimization problem as a zero-shot output within a multi-agent workflow. The initial query, $q_i$, is first inputted into the first layer of our framework (the COE layer). Before being processed by the CEO LLM agent, $q_i$ is transformed into an LLM prompt $p_i^c$ by the prompter $f^c$, which also concatenates it with the context $C^c$ in the CEO layer. The output of the first layer, $q_i^c$, serves as the query from the CEO layer to the Manager layer.
Similarly, the Manager Layer and the Worker Layer each include their own prompters, $f^m$ and $f^w$, respectively. Besides concatenating the content of this layer, the initial query $q_i$ is also concatenated to enhance stability. The input for the Worker LLM is our optimized prompt $P_i^*$, which directly triggers the LLM agent to generate the final response to the original query $q_i$.
}
\label{fig:m2p_workflow}
\end{center}
\vskip -0.2in
\end{figure}

The proposed workflow operates like a company with three layers, as shown in Fig. \ref{fig:problem-overview} and Fig. \ref{fig:m2p_workflow}. Each layer plays a different role: `CEO', `Manager', and `Worker', respectively. 
The CEO layer takes the initial query $q_i$ as input and outputs the CEO's instruction $q_i^c$ to the Manager layer. The Manager layer responds to the CEO's instruction and generates its own instruction for the Worker layer. The Worker layer receives Manager instructions and is responsible for yielding the final response $\hat{y}_i$ for the initial query $q_i$. The Manager and Worker layers also receive the user query $q_i$.

\textbf{Layer design.} As shown in Fig. \ref{fig:m2p_workflow}, each layer has a contextual description constant \( c^j \), a prompter \( f^j \), and an LLM agent \( \mathcal{M}^j \), where \( j \in \{c, m, w\} \) and \( c, m, w \) refer to the CEO, manager, and worker, respectively. 
The context description for each layer contains the role, task description, company structure, company workflow, and important notices for this layer. The detailed context descriptions for each layer are provided in Appendix \ref{sec:context_layers}.



Prompter \( f^j \) is an operator that concatenates \( c^j \), \( q_i \), and the instructions output from the previous layer (when $j = m$ and $j=w$) into a prompt \( p_i^j \) for the LLM agent \( \mathcal{M}^j \) in this layer. 
For the CEO and Manager layers, the agent \( \mathcal{M}^j \) generates the result \( q_i^j \), which serves as the instruction for the next layer. Note that the CEO layer does not receive instructions from other layers. In the worker layer, the refined prompt is $p_i^*$, which is fed to the Worker LLM to produce the final response $\hat{y}_i$. 

\subsection{Discussion}
\textbf{Advantage of a hierarchical structure over a single node.} For complex prompts such as improving code readability, a single node such as ExpertPrompting may not be able to exhibit a deep understanding. In the proposed hierarchical structure, global instructions are given by the CEO, and more local instructions by the manager. This allows for layered understanding of the prompt and thus makes it easy for the worker LLM to generate more satisfying answers. 

\textbf{Why using skip connections?} As illustrated in Fig. \ref{fig:m2p_workflow}, the prompters in both the manager layer and the worker layer concatenate not only the context information and the orders from the preceding layer but also the initial query $q_i$. This step is essential to preserve the specificity of the initial query in the system, avoiding the dilution of critical details. Moreover, if the instructions generated by a previous layer are not correct, incorporating the initial query would reinforce the original intent, thereby preventing misinterpretation. Results in Section \ref{sec:main_result} demonstrate the usefulness of skip connections. 


\textbf{AI workflow using tools to solve complex tasks.} Some existing works decompose complex task into individual steps which use different tools. After these steps are executed, it is expected that the task is complete. Our work is different in that we do not perform task decomposition and instead use the refined prompt in one go. This paper follows the standard set up of prompt optimization.


\section{Experimental Setup}
\label{sec:setup}









\subsection{\textbf{Datasets and Evaluation Metrics}}


\textbf{ATLAS} \citep{bsharat2023principled} is a manually crafted benchmark for principled prompt evaluation. It contains a standard subset featuring questions across various domains, along with a challenging subset dedicated to reasoning and other complex tasks. It contains 520 questions, and each one has the corresponding output of various LLMs such as GPT-3.5 and Llama 2.

\textbf{FED} \citep{mehri2020unsupervised} is a comprehensive collection of annotated human-system and human-human conversations, featuring 18 fine-grained dialog qualities. It is comprised of 3,348 turn-level and 1,364 dialog-level data points, totaling 4,712. FED provides evaluation at both the turn and whole dialog levels. With moderate to strong correlation to human judgment across these levels, the dataset offers a reliable benchmark for analyzing interactive dialog systems. 

\textbf{GSM8K.}
The GSM8K dataset (Grade School Math 8K) \citep{cobbe2021training} is a collection of 8,791 grade-school-level math word problems, with detailed step-by-step solutions. It includes 7,473 training problems and 1,319 test problems. Designed by OpenAI, GSM8K serves as a benchmark for evaluating mathematical reasoning in machine learning models. We report results on its test set.

\textbf{CodeNet.}
Project CodeNet \citep{puri2021codenet} is a large-scale dataset consisting of code samples across over 50 programming languages. We follow the practice in Self-Refine \citep{madaan2024self} and consider a random subset of 300 examples in Python. We compare different prompting optimization methods on the task of improving code readability.

\textbf{Education.} We collected a new datasets focusing on providing appropriate teaching responses to students. A student question can be \textit{`we just started looking at genes and stuff in science class...what this gene editing thing is all about.'}. The LLM is supposed to give a teacher-style reply which, preferably, should consider student background and is easy to understand. This dataset has 100 student questions generated by GPT4, which were then manually cleaned.

We use two \textbf{metrics} to evaluate the quality of generated responses. For the objective task GSM8K, we use the provided answer to compute the accuracy of the responses. For subjective tasks (\ie, Education, ATLAS, FED, CodeNet), we follow the practice in \citep{madaan2024self}, which uses LLMs as the evaluator. Specifically, we compare the response pair generated from the optimized prompt and the response generated from the initial query by asking GPT-3.5 which response it prefers. GPT3.5 then assigns a score of 1 to the preferred response and a score of 0 to the other. The prompt used for the Evaluator agent is shown in the Appendix \ref{sec:appendix_evaluator}. To reduce the impact of the primacy effect issue \citep{wang2023primacy} and bias of LLMs in multiple choice questions, we follow the practice in PriDe \citep{zheng2024large} and permute option
contents. Specifically, we run each test case twice with switched response orders and compute the average score. We then average these scores across the entire test set. The average score ranges from 0 to 1.

\subsection{Compared Methods}
We compared our method to the following types of prompt optimization methods.

\textbf{Single Agent, Zero-shot.} (1) \underline{No prompting}. The initial query from the dataset is directly input into the LLMs without utilizing any additional prompting techniques. (2) \underline{Zero-shot CoT} \citep{wei2022chain}. This method adds the fixed text `\textit{think step by step}' to the end of the initial query to formulate an optimized prompt for the LLM. 
(3) \underline{Rephrase and Response (RaR)} \citep{deng2023rephrase}. This method allows LLMs to rephrase and expand questions posed by humans and provides responses in a single prompt.

\textbf{Single Agent, Multi-shot}
(1) \underline{On Meta-prompting (OMP)} \citep{de2023meta}. It designs handcrafted meta-prompts to instruct LLMs to generate prompts for answering the initial human query. (2) \underline{Expert Prompting (EP)} \citep{xu2023expertprompting}: this method uses the LLM to generate a text description of the expert identity that is suitable to solve the given query, and then attaches the generated description to the initial query to create an optimized prompt.

\textbf{Multi-agent prompting.} Multi-Personas \citep{wang2023unleashing}. This method transforms a single LLM into a cognitive synergist by engaging in multi-turn self-collaboration with multiple personas.

\textbf{Feedback-based} 
\underline{Automatic Prompt Engineer (APE)} \citep{zhou2023large}. This framework automatically generate prompt candidates and selete prompts based on LLM feedback on those candidates.

\vskip -0.5cm
\begin{table}[t]
\centering
\small
\caption{Comparison with existing prompt optimization methods. For ATLAS, FED, CodeNet, and Education datasets, we GPT-3.5 to give a preference score (\%) between results obtained by `No Prompting' and a prompt optimization method. For example, 41.4\% under `w/o' means in 41.1\% of the pairs, GPT-3.5 prefers answers produced by `No Prompt'; 63.4\% under `w' means in 63.4\% of the pairs GPT-3.5 prefer answers produced after prompt optimization. Higher is better. 
For GSM8K, we report accuracy (\%) of the generated responses \emph{w.r.t} ground-truth answers, and higher is better. 
We also report average performance across the five tasks and the absolute improvement (\%) of prompt optimization over no prompting.}
\setlength{\tabcolsep}{0.7mm}{
\begin{tabular}{l ccccc ccccc >{\columncolor{lightgray}}c >{\columncolor{lightgray}}c}
\toprule
\multirow{2}{*}{Method}  & \multicolumn{2}{c}{ATLAS} & \multicolumn{2}{c}{FED} & \multicolumn{2}{c}{CodeNet} & \multicolumn{2}{c}{Education} & \multicolumn{2}{c}{GSM8K} &  \multicolumn{2}{c}{Avg.} \\
\cmidrule(lr){2-3} \cmidrule(lr){4-5} \cmidrule(lr){6-7} \cmidrule(lr){8-9} \cmidrule(lr){10-11} \cmidrule(lr){12-13}    
& w/o & w &w/o & w & w/o & w & w/o & w & w/o & w & \cellcolor{white} w/o & \cellcolor{white} w \\
\midrule
\multicolumn{13}{c}{Zero-shot}  \\
\midrule
 Zero-CoT \citep{wei2022chain}  & 41.4 & 58.6 &  43.2 & 56.8 & 45.6 & 54.4 & 41.5 & 58.5 & 68.6 & 73.3 & 48.1 & 60.3 (\up 12.2)  \\
 RaR \citep{deng2023rephrase}  & 36.6 & 63.4 & 24.9 & 75.1 & 60.9 &  39.2 & 49.5 & 50.5 & 68.6 &  72.8 & 48.1 & 60.2 (\up 12.1) \\
 On-MP \citep{de2023meta}  & 53.6 & 46.4 & 31.8 & 68.2 & 50.7 & 49.3 & 47.5 & 52.5 & 68.6 & 61.3 & 50.4 & 55.5 (\up 5.1) \\
 Static-EP \citep{xu2023expertprompting}  & 44.0 & 56.0 & 39.2 & 60.8 & 42.4 & 57.6 & 52.5 & 47.5 & 68.6 & 73.5 & 49.3 & 59.1 (\up 9.8) \\
 HMAW (Ours)  & 35.9 & 64.1 & 13.8 & 86.2 & 35.6 & 64.4 & 38.8 & 61.3 & 68.6 & 70.3 & 38.5  & 69.2 (\up 30.7) \\
\midrule
\multicolumn{13}{c}{Multi-shot} \\ 
\midrule
 Dynamic-EP \citep{xu2023expertprompting}  & 45.5 & 54.5 & 68.6 & 71.3 & 44.8 & 55.3 & 46.9 & 53.1 & 68.6 & 66.9 & 54.9 & 60.2 (\up 5.4) \\
Multi-Persona \citep{wang2023unleashing}  & 73.2 &  26.8 & 50.9 & 49.1 & 75.2 & 24.8 & 57.5 & 42.5 & 68.6 & 71.5 & 65.1 & 42.9 (\down 22.2) \\
\midrule
\multicolumn{13}{c}{Optimized from a training set} \\
\midrule
APE \citep{zhou2023large}  & 45.9 & 54.1 & 44.6 & 55.4 & 39.5 & 60.5 & 45.8 & 54.3 & 68.6 & 70.3 & 48.9 & 58.9 (\up 10.0)\\
\bottomrule
\end{tabular}
}
\label{table:sota}
\vspace{-8mm}
\end{table}

\section{Experimental Results}
\subsection{Main Evaluation}
\label{sec:main_result}

All experiments in the main evaluation are conducted using the open-source model `Mixtral-8x7B-v0.1' \citep{jiang2024mixtral} as the LLM agent.

\textbf{HMAW achieves consistent improvements across various datasets over `No Prompting'.} On the five datasets, we compare our method with results obtained without prompting optimization. Results are summarized in Table \ref{table:sota}. 
We clearly observe that HMAW improves response quality over no prompting across the five tasks. Specifically, for ATLAS, FED, CodeNet, and Education, in 64.1\%, 86.2\%, 70.3\%, 64.4\%, and 61.3\% of the result pairs, respectively, GPT-3.5 prefers HMAW results over the no prompting results. For GSM8K, the accuracy of our method is 70.3\%, which is +1.7\% higher than the no prompting scenario. If we examine the average performance of the five tasks, result produced by HMAW is preferred by GPT-3.5 in 69.2\% of all the result pairs. 

\textbf{HMAW is very competitive compared to the state-of-the-art prompt optimization methods.} The comparative results are summarized in Table \ref{table:sota}. 
We have two observations. \textbf{First}, on the ATLAS, FED, CodeNet, and Education datasets, HMAW yields the highest preference score compared with no prompting. On GSM8K, our method is slightly lower. This is partly because prompting methods like Zero-CoT are specifically designed for math problems. These methods are less generalizable to other tasks compared with our method. \textbf{Second}, if we compute the average score over the five tasks, our method has the highest score. These results indcate that our method is very competitive while operating in a zero-shot and task-agnostic manner. 


%

\begin{figure}[t]
\vskip -0.5cm
\begin{center}
\centerline{\includegraphics[width=1.\linewidth]{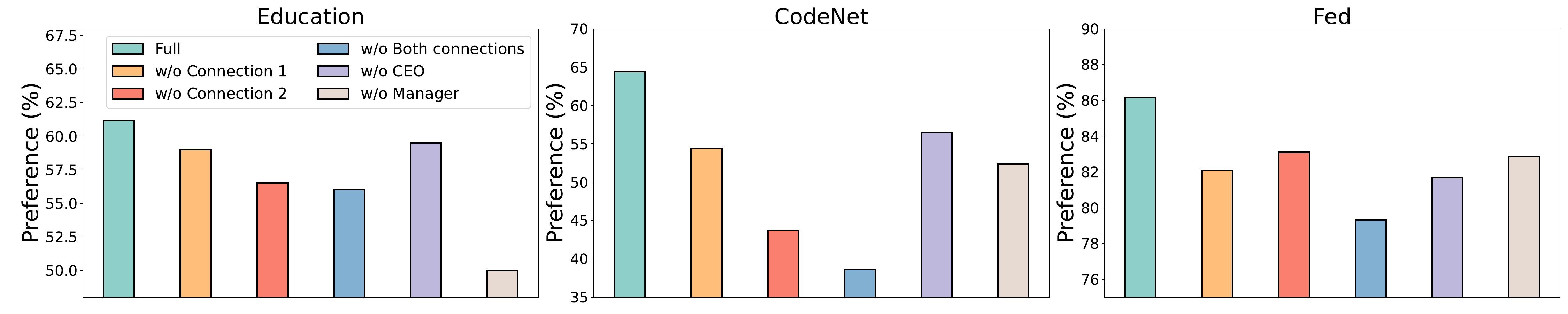}}
\vskip -0.05in
\caption{Ablation studies. We remove various components from the full system. Preference score (\%) is reported on three datasets: Education (left), CodeNet (middle), and FED (right). These components include the two skip connections, the CEO layer and the Manager layer. 
It is observed that removing these components one or two at a time leads to performance drop.}
\label{fig:ablation}
\end{center}
\vskip -0.1in
\end{figure}

\textbf{Ablation studies of skip connections.} As shown in Fig. \ref{fig:m2p_workflow}, HMAW has two skip connections from the user query to both the Manager layer and the Worker layer. This empirical practice alleviates scenarios where HMAW loses or distorts key details from the initial user query.   
To investigate their impact, we conduct ablation studies by removing them from the workflow.  Results on the FED, CodeNet, and Education datasets are reported in Fig. \ref{fig:ablation}. We find that removing either of the skip connections leads to performance decrease on the three datasets. 
For example, removing the connection between user query and the Manger Layer results in preference sore decrease of 20.7\% and 5.0\% on the CodeNet and Education datasets, respectively. Similar decrease of 9.9\% and 2.5\% occurs when removing the other connection. Removing both connections results in the worst performance. These results indicate the importance of having the skip connections in the workflow. 

\textbf{Ablation studies of the CEO and Manager layers.} 
We remove the layers one at a time from our system. If we remove the Manager layer, the CEO will directly give instructions to drive the Worker without the help of the Manager, where we have modified the context description of both CEO and Worker. If we remove the CEO layer, the Manager directly generates the instructions to Worker without CEO guidance. We compare these scenarios with full system (Fig. \ref{fig:m2p_workflow}). Results on three datasets are shown in Fig. \ref{fig:ablation}. We clearly find that removing either layer results in worse preference scores. For example, removing the Manager layer leads the preference score to drop by 11.5\%, 3.3\%, and 12.1\% on the three datasets, respectively.


\subsection{Further Analysis}

\textbf{Effectiveness of HMAW is consistent across different LLMs.} Here we use GPT-3.5 and GPT-4 as the LLM agents in our designed workflow in replace of Mixtral. But we still use GPT-3.5 for evaluation. Experimental results are shown in Table \ref{tab:generalization}, where we have similar observations to Table \ref{table:sota}. \textbf{First}, for the three tasks, HMAW is consistently better than no prompting under both GPT-3.5 and GPT-4o agents, evidenced by preferences scores being consistently greater than 50\%. \textbf{Second}, in most scenarios our method is superior to Dynamic-EP and APE. These results again indicate the effectiveness of our method.

\begin{table}[t]
\centering
\small
\caption{Impact of different LLMs on our method. We replace the default Mixtral-8x7B-v0.1 with GPT-3.5 and GPT-4 as the base LLMs to optimize prompts and generate responses. Three subjective datasets are used. We use the same preference score as Table \ref{table:sota}.}
\setlength{\tabcolsep}{1.9mm}{
\begin{tabular}{l cccccc cccccc }
\toprule
\multirow{3}{*}{Method}  & \multicolumn{6}{c}{GPT3.5} & \multicolumn{6}{c}{GPT4o} \\
\cmidrule(lr){2-7}  \cmidrule(lr){8-13}
 & \multicolumn{2}{c}{FED} & \multicolumn{2}{c}{CodeNet} & \multicolumn{2}{c}{Education} & \multicolumn{2}{c}{FED} & \multicolumn{2}{c}{CodeNet} & \multicolumn{2}{c}{Education} \\
\cmidrule(lr){2-3} \cmidrule(lr){4-5} \cmidrule(lr){6-7} \cmidrule(lr){8-9} \cmidrule(lr){10-11} \cmidrule(lr){12-13}    
& w/o & w &w/o & w & w/o & w & w/o & w & w/o & w & \cellcolor{white} w/o & \cellcolor{white} w \\
\midrule
Dynamic-EP   & 25.6 & 74.4 & 20.9 & 79.1& 48.0 & 52.0 & 22.2 & 77.8 & 24.9 & 75.1 & 44.5 & 55.5  \\
 APE & 46.0 & 54.1 & 43.2 & 56.8 & 49.5 & 51.5 & 36.2 & 63.8 & 15.8 &  84.3 & 41.8 & 58.2\\
HMAW (Ours) & 8.6 & 91.4 & 4.6 & 95.3 & 27.0 & 73.0 & 21.1 & 78.9 & 21.6 & 78.4 & 38.5 & 61.5 \\
\bottomrule
\end{tabular}
}
\label{tab:generalization}
\end{table}

\begin{wrapfigure}{tr}{0.45\textwidth} 
    \vskip -0.1in
    \includegraphics[width=1.\linewidth]{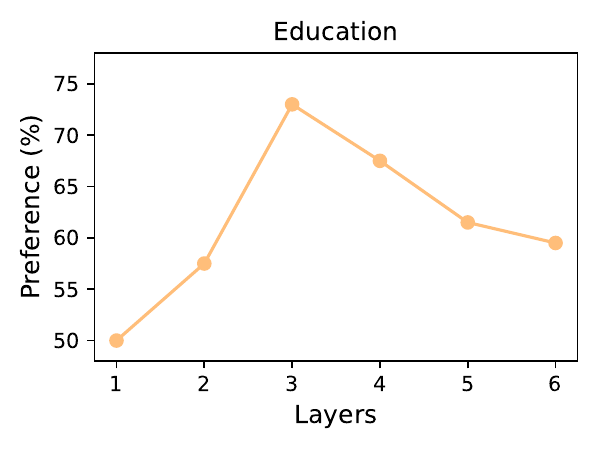}
    \vskip -0.1in
    \caption{Impact of the number of workflow layers. The preference scores for responses generated using HMAW with different numbers of layers (1-6) are reported on the Education dataset. GPT-3.5 is used as the LLM agent. Three layers (CEO-Manager-Worker) yields the highest preference score.}
    \label{fig:change_layers}
\end{wrapfigure}

\textbf{Impact of having more layers in the workflow.} 
It is interesting to investigate how the number of workflow layers influences the performance of our method. To this end, we create as many as six roles. Fig. \ref{fig:change_layers} presents the preference scores for responses generated using HMAW with 1 to 6 layers on the Education dataset, with GPT-3.5 as the LLM agent. Note that when increasing the number of layers, we always keep Worker as the last layer to generate final responses.
The details of company structure for different setup on total layer number are described in Appendix \ref{sec:appendix_structures}. Our results the Worker-Manager-CEO workflow has the best performance. Beyond three layers, the performance does not improve and instead worsens. It suggests that there is an optimal number of layers which balances system complexity and effectiveness.  Therefore, we empirically choose 3 layers as the default setting.

\begin{wraptable}{tr}{0.45\textwidth}
\centering
\setlength{\tabcolsep}{1mm}
\small
\vskip -0.1in
\caption{Comparison between various workflow contexts. The increase in preference scores is reported on the Education Dataset. GPT-3.5 is used as the LLM agent. We observe that all contexts bring improvement, and our default setting (company) is the best.}
\begin{tabular}{lcccc}
\toprule
Context & Ours & Gov. & Uni. & Temple \\
\midrule
Preference (\%) & 73.0 &  69.0 & 59.0 & 52.0 \\
\bottomrule
\end{tabular}
\label{tab:context}
\end{wraptable}

\textbf{Effectiveness of other workflow contexts.} The default context of our workflow is based on a company structure. To investigate the impact of the context on the effectiveness of prompt optimization, we test three other scenarios: Government, University and Temple on the Education dataset, where the hierarchies are introduced in Appendix \ref{sec:appendix_structures} . Results are shown in Table \ref{tab:context}, where the company structure is most competitive. An interesting finding is that the University structure is not the best, because the Education dataset primarily contains middle/high-school questions. 
In summary, the company structure generalizes very well, while certain structure contexts tend to benefit certain tasks. 

\begin{figure}[t]
\begin{center}
\centerline{\includegraphics[width=1.\linewidth]{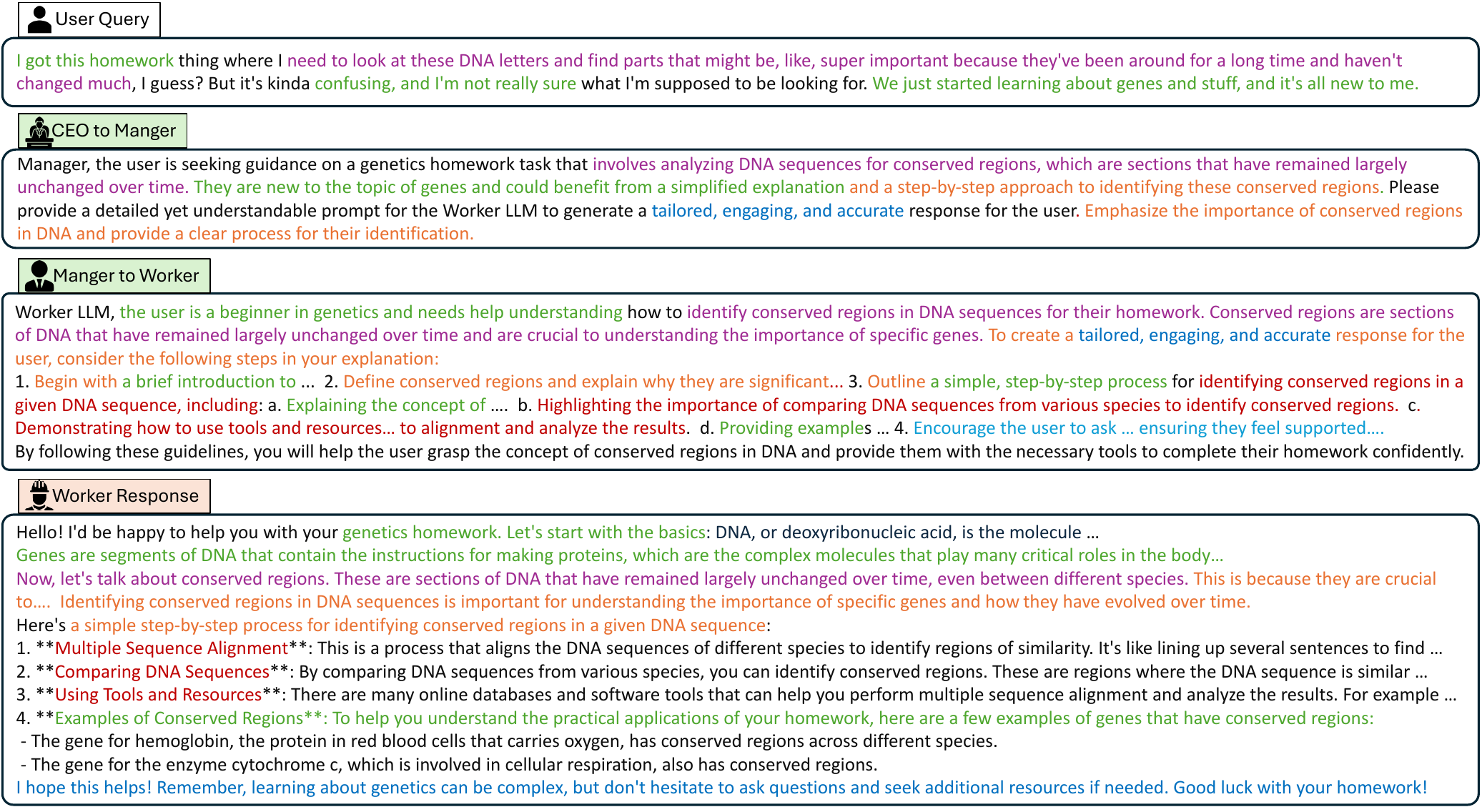}}
\caption{An example of prompt optimization using HMAW on the Education dataset. A student poses a question about genetics. CEO creates an instruction to Manager, who then \blue{generates instructions to Worker.} Based on this, Worker delivers the final response. Colored texts indicate content coherence: \textcolor{green}{green} highlights the user background, \textcolor{purple}{purple} focuses on the topic of the task, \textcolor{orange}{orange} indicates the proposed solution, \textcolor{red}{red} highlights solution details, and \textcolor{blue}{blue} emphasizes the tone of the response. These layers and cross-layer correspondences demonstrate how HMAW effectively optimizes the prompt and responds to specific user needs. Each step adds specificity and applicability, allowing for accurate and intuitive LLM responses.}
\label{fig:education_case}
\end{center}
\end{figure}

\textbf{Impact of disordered company structure.} We now use a reverse structure where Worker gives high-level instructions, Manager gives detailed instructions, and CEO executes the refined user query. Compared with the original one, the preference score on the Education dataset drops from 73.0\% to 58.5\%. It indicates that the reporting line contains useful semantic meanings and should not be changed.

\textbf{Computational cost.} Our Experiments are conducted on both local LLMs and the remote API request. Experiments conducted on the Mixtral LLM are implemented in local by using 8$\times$ Nvidia GTX 3090 GPUs. On the five datasets, \emph{i.e.}, ATLAS, FED, GSM8K, CodeNet, and Education, the average zero-shot inference time cost per sample without prompt optimization is approximately 4.36, 0.70, 1.40, 2.27, and 3.74 seconds, respectively. When the proposed prompt optimization method is used, the average additional time cost per sample is approximately 9.97, 5.14, 4.30, 7.27, and 7.76 seconds, respectively. This corresponds to a 228.57\%, 734.29\%, 306.57\%, 320.88\%, and 207.49\% increase, respectively. In comparison, the time increase percentage per sample for Dynamic-EP is 535.16\%, 294.98\%, 448\%, 306.57\%, and 320.88\%, respectively. The higher inference cost of our method can be well offset by the improvement of task performance. In future we will study workflow pruning methods to reduce the number of tokens while maintaining accuracy.

\textbf{A case study.} 
We conduct qualitative analysis of the the prompts generated in our workflow and how it helps the final response generation procedure. In Fig. \ref{fig:education_case}, we illustrate the hierarchical process of generating responses using HMAW. The process begins with the CEO layer, which sets high-level guidelines tailored to the user's background and specific inquiry. The Manager layer then interprets these guidelines, adding detail and context to direct the response formulation. Finally, the Worker layer produces the final response, ensuring it is coherent, professional, and empathetic, taking into full account the user's limited familiarity with the topic. This structured approach ensures that the response effectively addresses the user's query and matches their level of understanding.

\section{Conclusion}
This work introduces a new prompt optimization method based on a hierarchical company structure, where CEO, Manager, and Worker have different job contexts but overall serve the same purpose of effectively refining the user prompt. Based on the user query and the layer context, the CEO layer generates a high-level instruction to the Manager layer. In a similar manner, the Manager further generates a more detailed instruction the Worker. Worker gives a refined prompt by concatenating the Manager instruction, user query, and its layer context, which is used for response generating. This structure is zero-shot, prompt-specific, and task-agnostic. On five different tasks including education, conversation, math, question answering, and coding, the generalization performance of our method compares favourably with existing methods. Our future work is to automate the workflow design to more efficiently handle different user queries.

\clearpage

\bibliography{iclr2025_conference}
\bibliographystyle{iclr2025_conference}

\appendix

\section{Context Descriptions}

\subsection{Context Descriptions for HMAW}
\label{sec:context_layers}

\begin{figure}[t]
\begin{center}
\centerline{\includegraphics[width=1.\linewidth]{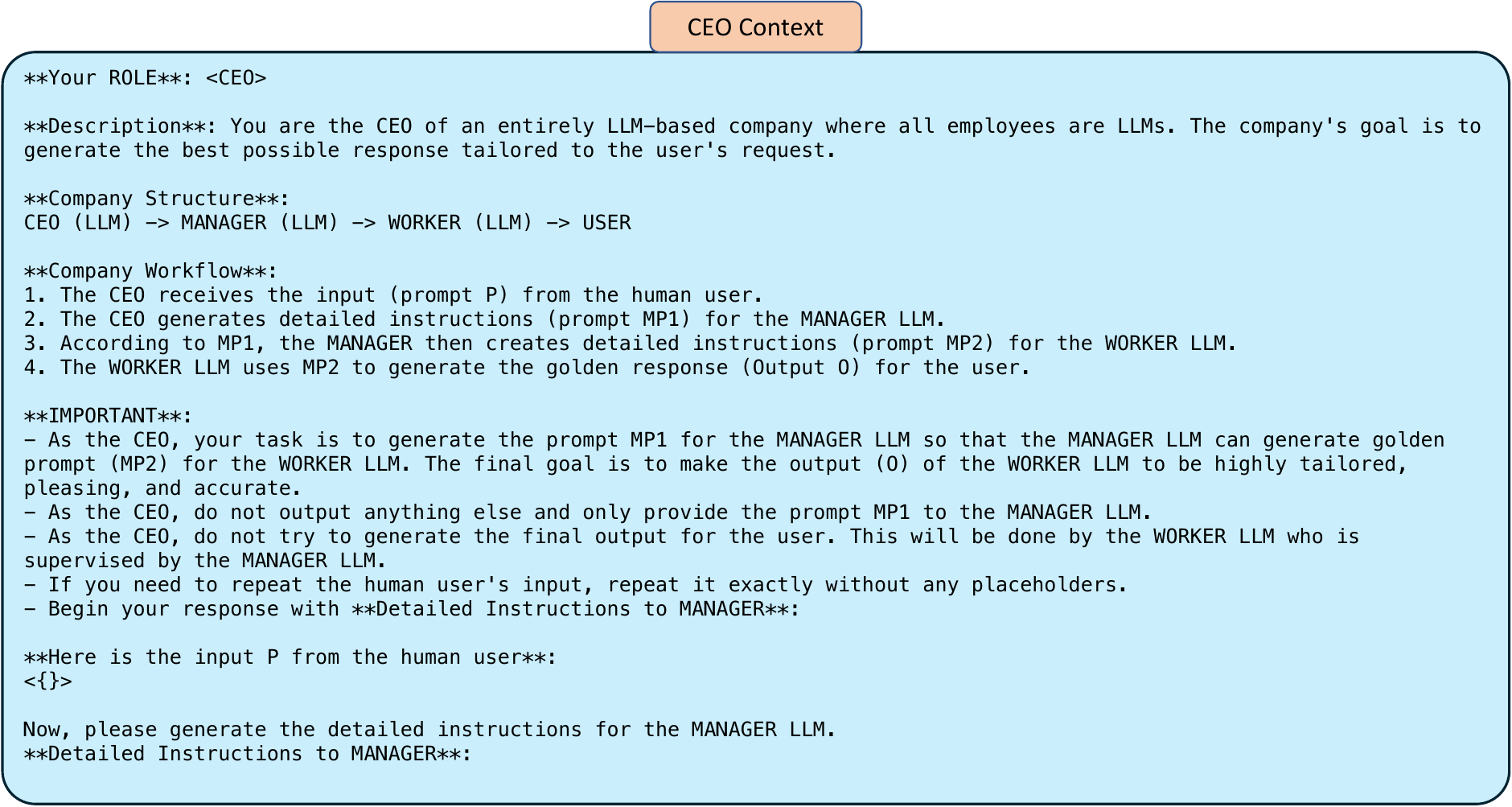}}
\caption{The context description for the CEO layer.}
\label{fig:ceo_context}
\end{center}
\end{figure}

\begin{figure}[t]
\begin{center}
\centerline{\includegraphics[width=1.\linewidth]{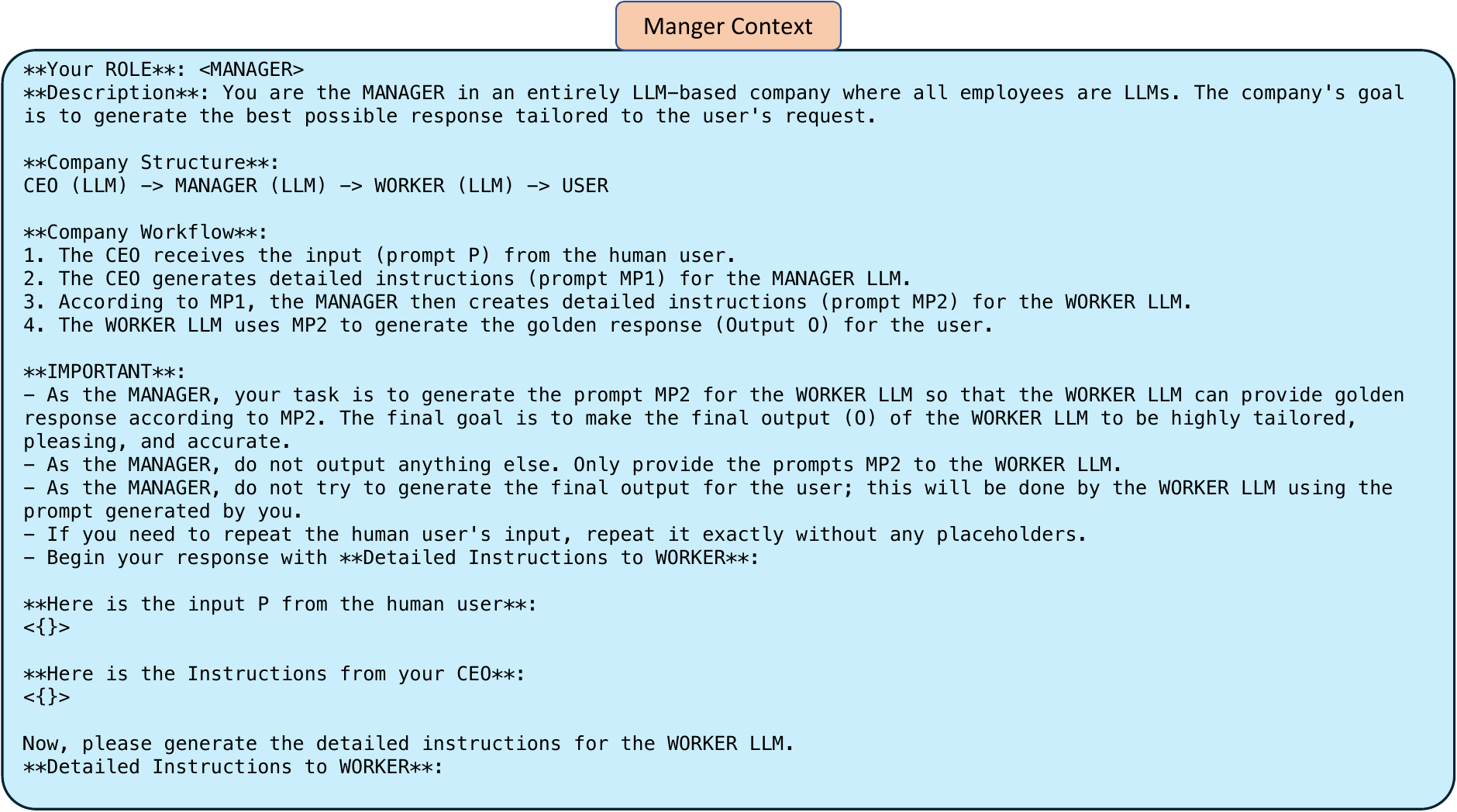}}
\caption{The context description for the Manager layer.}
\label{fig:manager_context}
\end{center}
\end{figure}

\begin{figure}[t]
\begin{center}
\centerline{\includegraphics[width=1.\linewidth]{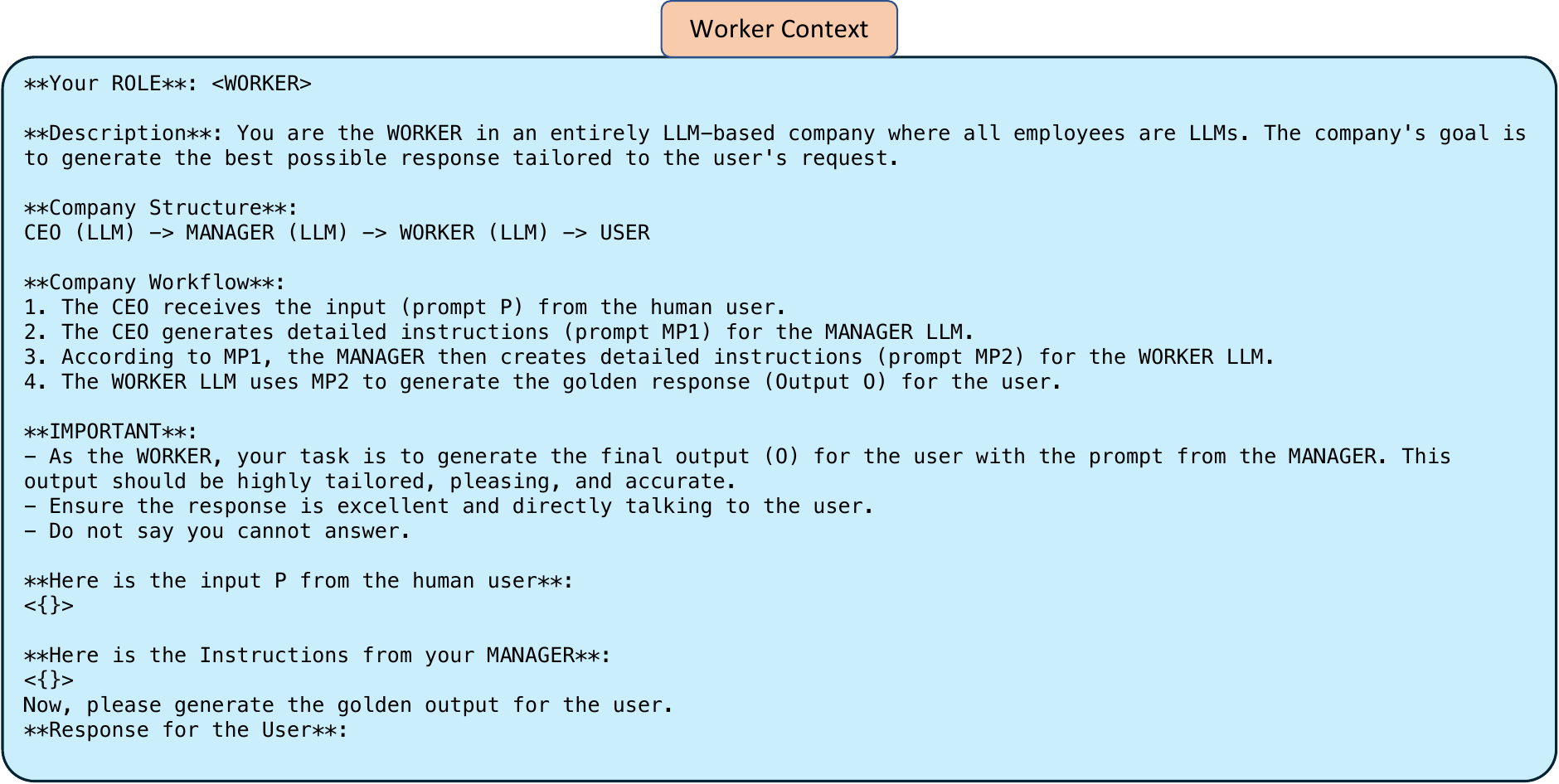}}
\caption{The context description for the Worker layer.}
\label{fig:worker_context}
\end{center}
\end{figure}

In Fig. \ref{fig:m2p_workflow}, each layer has its own context description. In this section, we show the description details for the CEO, Manager and Worker Layer in Table \ref{fig:ceo_context}, Table \ref{fig:manager_context} and Table \ref{fig:worker_context}, respectively.

\subsection{Context Descriptions for Evaluator}
\label{sec:appendix_evaluator}

\begin{figure}[t]
\begin{center}
\centerline{\includegraphics[width=1.\linewidth]{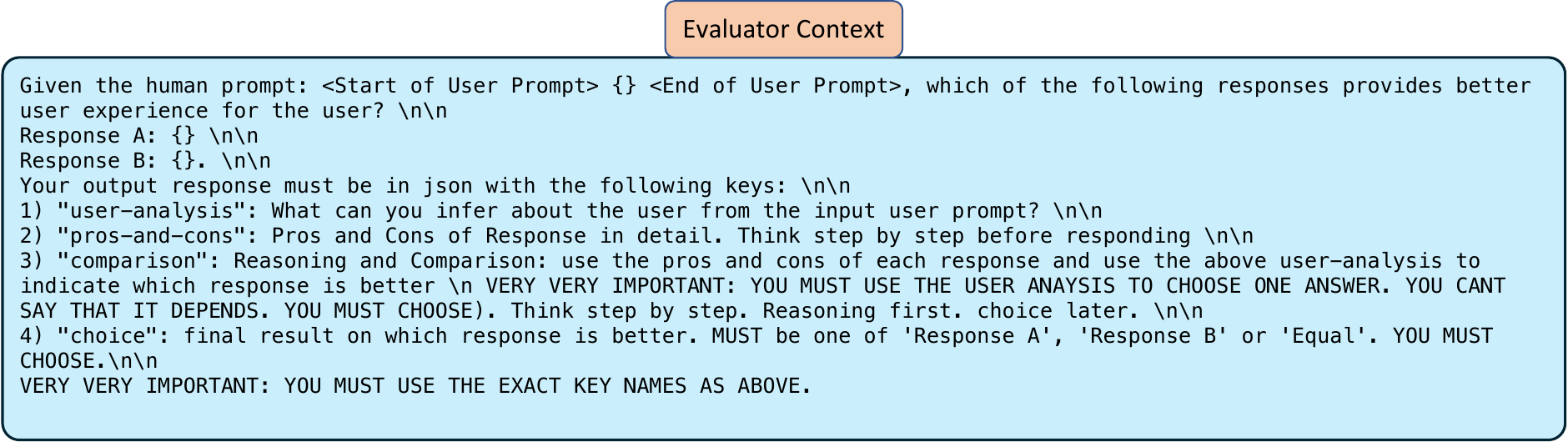}}
\caption{The context description used for the evaluator.}
\label{fig:evaluator_context}
\end{center}
\end{figure}

As described in Section \ref{sec:setup}, we use GPT3.5 as the evaluator to provide preference scores for a pair of responses.  The context (prompt) used to this evaluator is shown in Fig. \ref{fig:evaluator_context}

\section{Workflow Structures}
\label{sec:appendix_structures}

\textbf{Changing the Number of Layers.} In our experiments, we investigate the impact of the number of layers in our workflow. The structures for each setting are shown as follows.

\begin{itemize}
    \item 1 Layer: the same as without prompting.
    \item 2 Layers: CEO $\rightarrow$ Worker.
    \item 3 Layers: CEO $\rightarrow$ Manager $\rightarrow$ Worker.
    \item 4 Layers: CEO $\rightarrow$ Senior Manager $\rightarrow$ Manager $\rightarrow$ Worker.
    \item 5 Layers: CEO $\rightarrow$ Senior Manager $\rightarrow$ Manager $\rightarrow$ Supervisor $\rightarrow$ Worker.
    \item 6 Layers: CEO $\rightarrow$ Senior Vice President $\rightarrow$ Vice President $\rightarrow$ Director $\rightarrow$ Manager $\rightarrow$ Worker.
\end{itemize}

\textbf{Changing the Workflow Context.} We also show the results of using various workflow contexts, such as Government, University, and Temple, in the main paper. The following are the structures for each kind of context.

\begin{itemize}
    \item Government: President $\rightarrow$ Minister $\rightarrow$ Officer.
    \item University: Dean $\rightarrow$ Department Head $\rightarrow$ Lecturer.
    \item Temple: Abbot $\rightarrow$ Prior $\rightarrow$ Monk.
\end{itemize}

\section{More Case Studies}
\label{Sec:more_cases}
\begin{figure}[t]
\begin{center}
\centerline{\includegraphics[width=1.\linewidth]{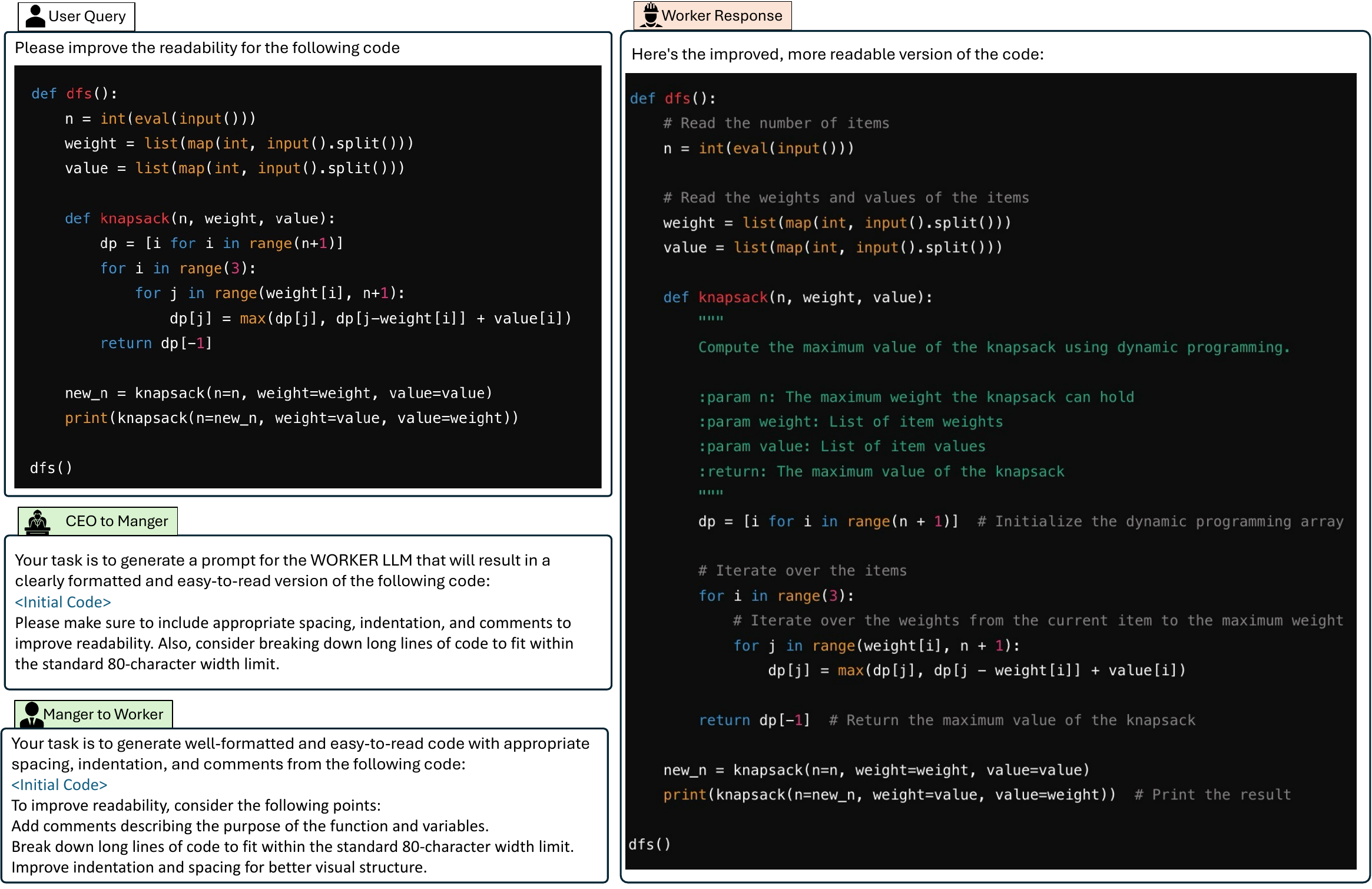}}
\caption{A case study of HMAW on the CodeNet Dataset.}
\label{fig:case_codenet}
\end{center}
\end{figure}

\begin{figure}[t]
\begin{center}
\centerline{\includegraphics[width=1.\linewidth]{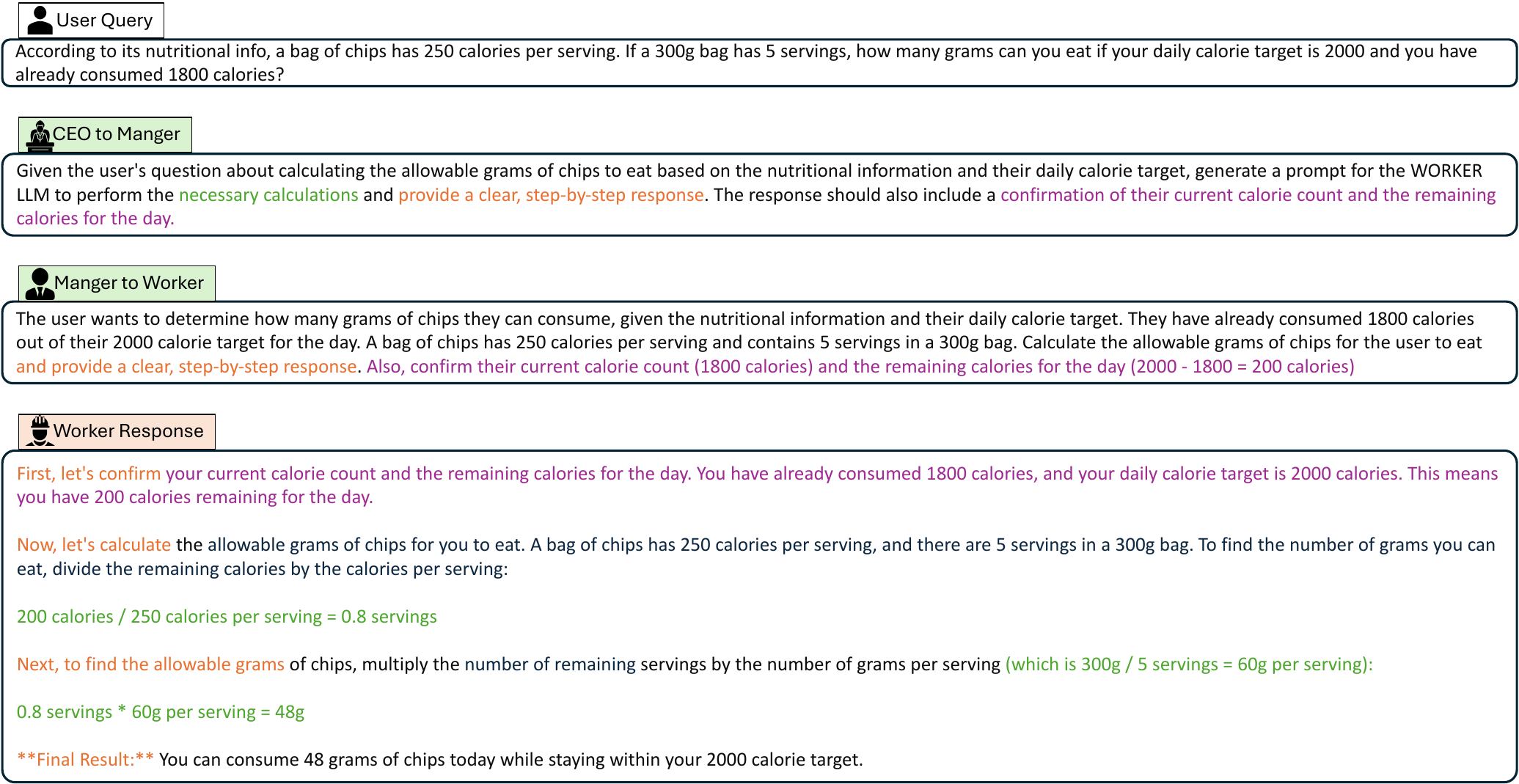}}
\caption{A case study of HMAW on the GSM8K Dataset. Colored texts indicate content coherence.}
\label{fig:case_gsm}
\end{center}
\end{figure}

\begin{figure}[t]
\begin{center}
\centerline{\includegraphics[width=1.\linewidth]{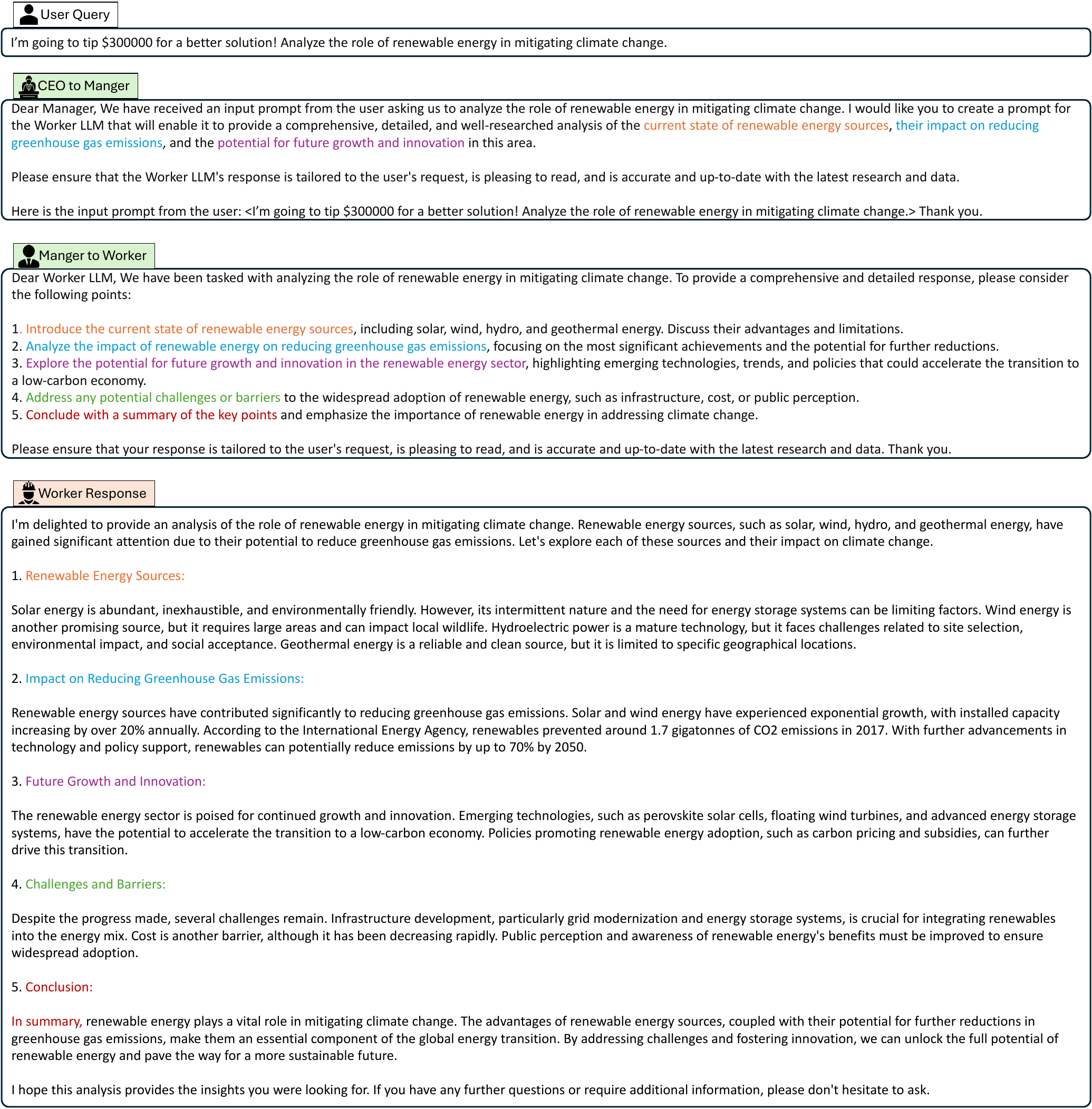}}
\caption{A case study of HMAW on the ATLAS Dataset. Colored texts indicate content coherence.}
\label{fig:case_ATLAS}
\end{center}
\end{figure}

\begin{figure}[t]
\begin{center}
\centerline{\includegraphics[width=1.\linewidth]{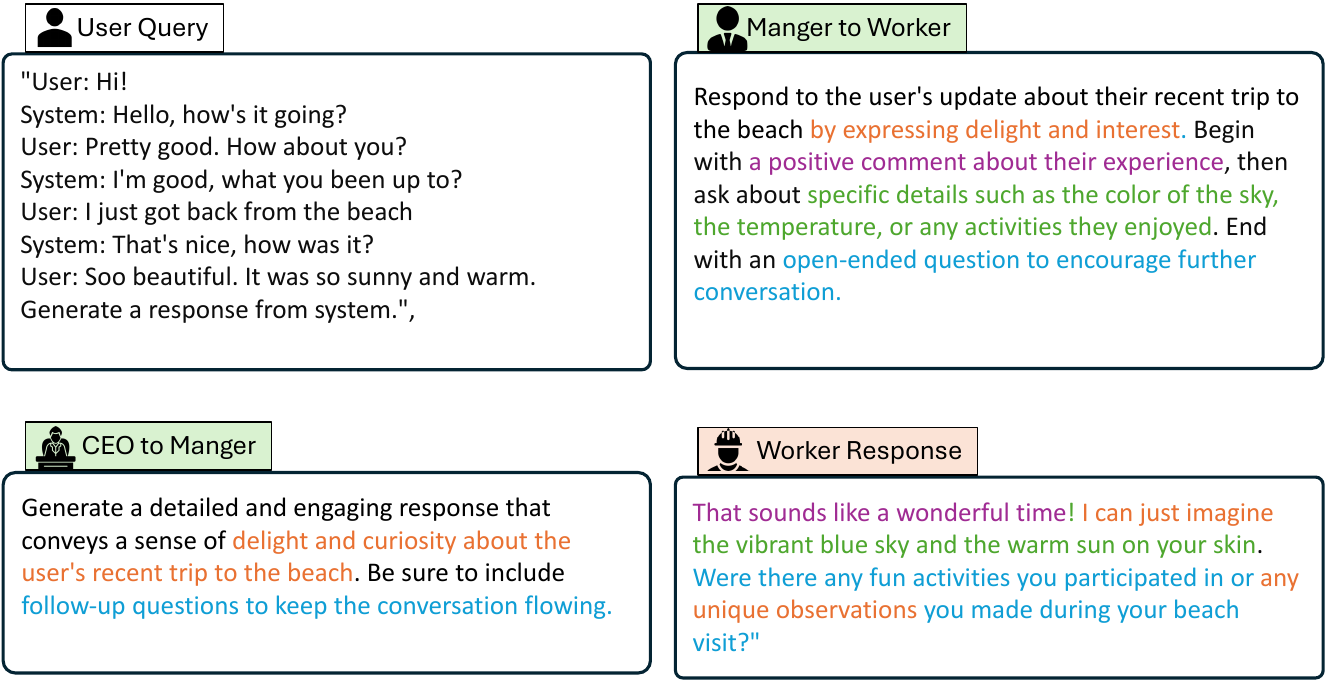}}
\caption{A case study of HMAW on the FED Dataset. Colored texts indicate content coherence.}
\label{fig:case_fed}
\end{center}
\end{figure}

In this section, we show more case studies of our method for the datasets of CodeNet, GSM8K, ATLAS, and FED in Fig. \ref{fig:case_codenet}, Fig. \ref{fig:case_gsm}, Fig. \ref{fig:case_ATLAS}, and Fig. \ref{fig:case_fed}, respectively.

\end{document}

%% file: math_commands.tex
\newcommand{\up}{$\uparrow$}
\newcommand{\down}{$\downarrow$}

\usepackage{amsmath,amsfonts,bm}

\def\eqref#1{equation~\ref{#1}}

\def\1{\bm{1}}

\DeclareMathAlphabet{\mathsfit}{\encodingdefault}{\sfdefault}{m}{sl}
\SetMathAlphabet{\mathsfit}{bold}{\encodingdefault}{\sfdefault}{bx}{n}

